\documentclass[10pt,twocolumn,letterpaper]{article}
\pdfoutput=1 
\usepackage{cvpr}
\usepackage{times}
\usepackage{epsfig}
\usepackage{graphicx}
\usepackage{amsmath}
\usepackage{amssymb}

\usepackage[breaklinks=true,bookmarks=false]{hyperref}

\cvprfinalcopy 

\def\cvprPaperID{****} 

\ifcvprfinal\fi
\begin{document}

\title{Multi-Scale Body-Part Mask Guided Attention for Person Re-identification}

\author{Honglong Cai  \qquad Zhiguan Wang \qquad Jinxing Cheng\\
Suning R\&D Center USA\\
{\tt\small \{honglong.cai, doris.wang,  jim.cheng\}@ussuning.com}
}

\maketitle
\thispagestyle{empty}

\begin{abstract}
Person re-identification becomes a more and more important task due to its wide applications. In practice, person re-identification still remains challenging due to the variation of person pose, different lighting, occlusion, misalignment, background clutter, etc. In this paper, we propose a multi-scale body-part mask guided attention network (MMGA), which jointly learns whole-body and part-body attention to help extract global and local features simultaneously. In MMGA, body-part masks are used to guide the training of corresponding attention. Experiments show that our proposed method can reduce the negative influence of variation of person pose, misalignment and background clutter. Our method achieves rank-1/mAP of 95.0\%/87.2\% on the Market1501 dataset, 89.5\%/78.1\% on the DukeMTMC-reID dataset, outperforming current state-of-the-art methods.
\end{abstract}
\section{Introduction}
\begin{figure}
    \includegraphics[width=0.5\textwidth]{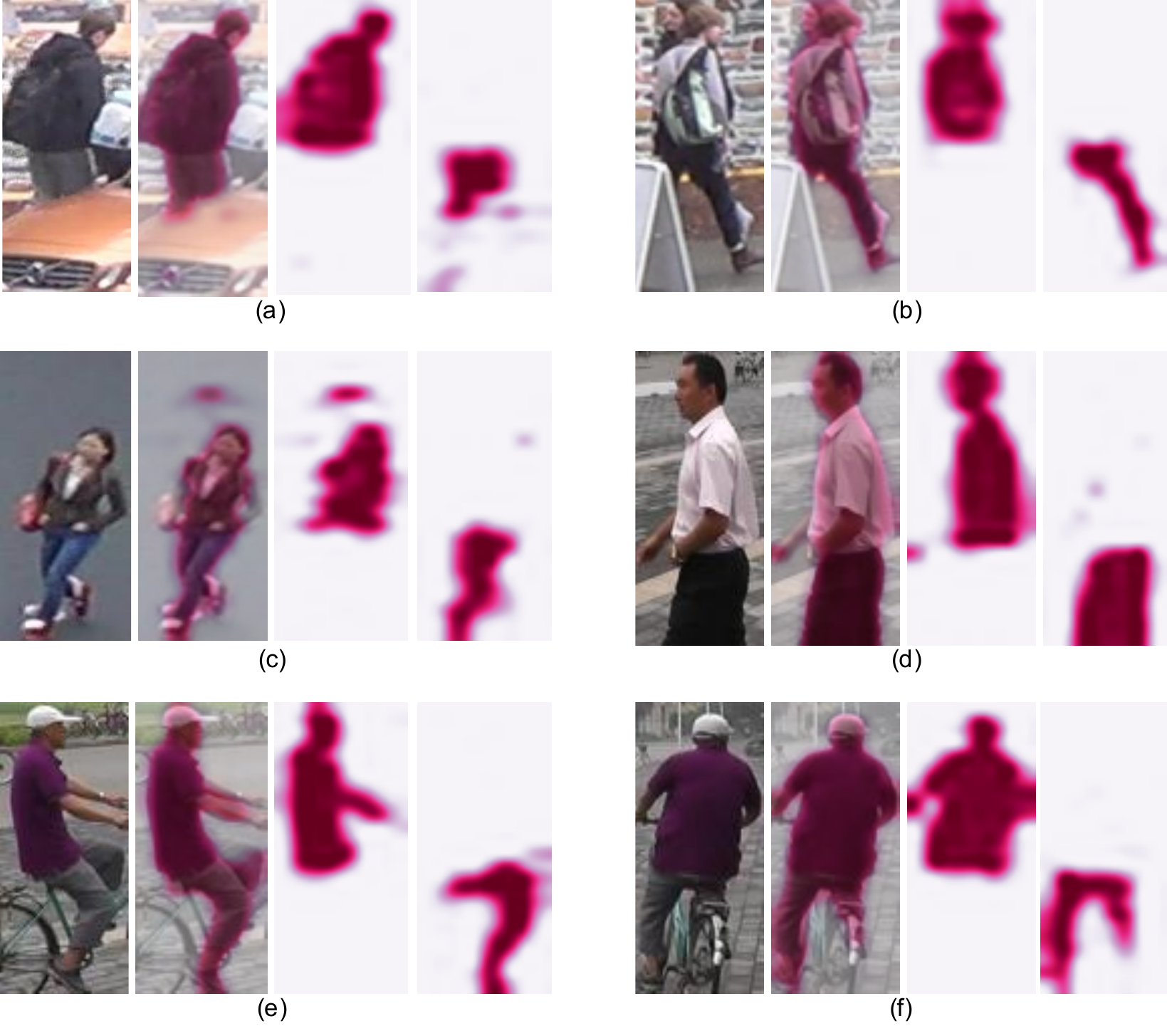}
    \caption{Examples of challenges in person re-identification and how our attention mechanism can handle the challenges. (a-b) occlusions, (c-d) inaccurate bounding-box detection, (e-f) variation of pose. The second to forth images in each group are the global attention map in original image, upper-body attention map and bottom-body attention map generated by our proposed MMGA network.}
    \label{fig:Figure1}
\end{figure}

Person re-identification (re-ID) aims at identifying the presence of same person in different cameras with different backgrounds, poses and positions. It is still a challenging task due to large variations on persons like pose, occlusion, clothes, background clutter and detection failure which are shown in Figure~\ref{fig:Figure1}. 

Low-level features like colors, shapes, contours and local descriptors are used to train traditional re-ID models with low accuracy \cite{farenzena2010person,hamdoun2008person}. Nowadays, with the fast development of deep neural networks, deep features of human image learned through convolutional neural network (CNN) is demonstrated to have better discrimination and robustness to represent the image, which has made significant improvement on the re-ID problem \cite{cheng2016person,lin2017improving,matsukawa2016person,varior2016gated}. The features learned from deep learning network should capture the most salient clues that can represent identities of different persons. However, most of the existing deep learning methods learn features from the whole image that contains not only the human body parts, but also the background regions \cite{xiao2016learning,wang2016joint}. The background regions containing clutter and occlusions may lead to a misalignment problem . To address this issue, some recent works \cite{pcb,mgn,alignedreid,cheng2016person,varior2016siamese,contextaware} show that locating the significant body parts and learning the discriminative features from these informative regions can reduce the negative effects of clutter and occlusions, and thus improve the re-ID accuracy. 

Visual attention has shown its success in re-ID tasks \cite{attentiondriven,hacnn, maskguided, diversity}, as the mechanism conforms to the human visual system that a whole image is not likely to be processed in its entirety at once, but only the salient parts of the whole visual space are focused when and where needed. Visual attention module can help to extract dynamic features from salient parts mostly like human body parts in a image by guiding the learning towards informative image regions \cite{maskguided}. Given the human body information, attention maps where regions of interest are presented have much stronger responses on body region compared with background regions \cite{hacnn,attentiondriven}. Inspired by this, whole human body mask has been used to guide the attention model training \cite{maskguided}. The whole human body is segmented from the background region and then attention network is guided by this binary mask. However, learning global attention through the whole human body mask may suppress local informative body part regions that have stronger responses. 

To relieve this dilemma, we introduce a multi-scale body-part mask guided attention network. We split body masks into upper-body masks and bottom-body masks and use them to guide the training of upper-body attention and bottom-body attention respectively. As shown in Figure~\ref{fig:Figure1}, our proposed network can learn whole-body attention, upper-body attention and bottom-body attention. Moreover, comparing to \cite{maskguided} , which needs mask in both training and inference phase, our proposed method needs mask only in training phase, and thus saves significant time in inference phase.

Our experiments indicate that multi-scale body-part mask guided attention network can significantly improve the accuracy of person re-identification and it still has space for improvement. We conclude our paper with qualitative results and demonstrate the potential of the method.

The contributions of our work can be summarized as follows:
\begin{itemize}
    \item We propose a mask guided attention method to address person re-identification problem. In our proposed method, mask is used to guide attention training only in the training phase. We don't need mask in inference phase which makes our method particularly efficient.
    \item We creatively use the masks of different parts of body to guide attention learning in training phase. In this paper, we separate person body into upper part and bottom part and use them to guide the training of upper and bottom attention respectively.
\end{itemize}

\section{Related Work}
This section provides an overview of closely related work in deep person re-identification. Deep learning has been successfully used to improve the performance of person re-identification. Both verification and identification models are applied in deep person re-identification \cite{multiscale, deepreid, varior2016gated, deepperson}.


\noindent\textbf{Part-Based Model:} Recently, a number of new methods have been designed to capture richer and finer visual cues by jointly learning from not only whole-body images but also body-part images, which pushed the re-ID performances to a new level. According to different partition strategies, the part-body based methods can be classified into three categories: 1) Human body parts are generated through predefined fixed-height horizontal stripes. In work \cite{alignedreid,pcb, horizontal, mgn}, they both equally slice the feature maps of input image in vertical orientation.  2) The human body-part regions are detected and the local features are generated jointly. \cite{Yao2019DeepRL} proposed a method that can estimate body parts in a feature space through ROI pooling, while the local features can be generated at the same time. 3) Human body can be divided into different parts according to the keypoints detected by off-the-shelf pose estimation model. In work \cite{attentiondriven}, 18 human body joints are obtained through pose estimation model and 5 body regions are defined according to the body joints. These methods all focus on the part partition scheme for the local feature extraction network. In our proposed method, we generate the body-part mask and use the local mask to guide our attention model, as human vision may not focus on the whole human body but only on some body parts in a image.

\noindent\textbf{Attention mechanism in Re-ID:}
One difficulty person re-identification suffering is misalignment, as the bounding-boxes detected by the detection algorithm may not be accurate enough, for example only partial profile of pedestrian may appear at the corner of the image. To overcome this issue, attention mechanism is proposed to aid the network to learn where to 'look' at. Attention mechanism plays a more and more important role in computer vision field \cite{hacnn, mancs, Liu2017HydraPlusNetAD, nonlocal, selfattention, diversity, attentiondriven, residualattention, sca}. In \cite{hacnn}, a harmonious attention model is introduced to combine both soft and hard attention mechanism. A multi-directional attention model is proposed in \cite{Liu2017HydraPlusNetAD}, which can extracts attentive features through masking different levels of features with attention map. Given that most of the attention based methods generate global attention through whole-body images, the local attention learning from each body part is ignored, which may lead to suboptimal performance when person images suffer from large pose variations, misalignment, local occlusion, etc. To address this issue, \cite{diversity} proposes a spatiotemporal attention model which can automatically discover a diverse set of distinctive body parts. In our proposed model, the local attention can be learned through local mask, thus the re-ID performance can be further improved.

\noindent\textbf{Mask mechanism in Re-ID:} Human body mask obtained from image segmentation models can be used to handle the background clutter problem. With deep learning based image segmentation algorithms including Mask RCNN \cite{He2017MaskR}, JPPNet \cite{jppnetj}, Dense Pose \cite{densepose}, etc., human body mask can be extracted well and the background region can be almost removed. However, there are only a few works \cite{maskguided,Chen2018PersonSV,spreid} introducing semantic segmentation into re-ID task. This scarcity is due to large computational complexity involved in semantic segmentation for human mask. In our work, we just utilize the mask to guide the training of our attention model so that the mask is only needed in training phase. After we get the learning metric, mask is no longer needed to extract features which is time-saving compared with other mask based re-ID algorithms. 

\section{Our Proposed Method}
\begin{figure}
    \includegraphics[width=0.5\textwidth]{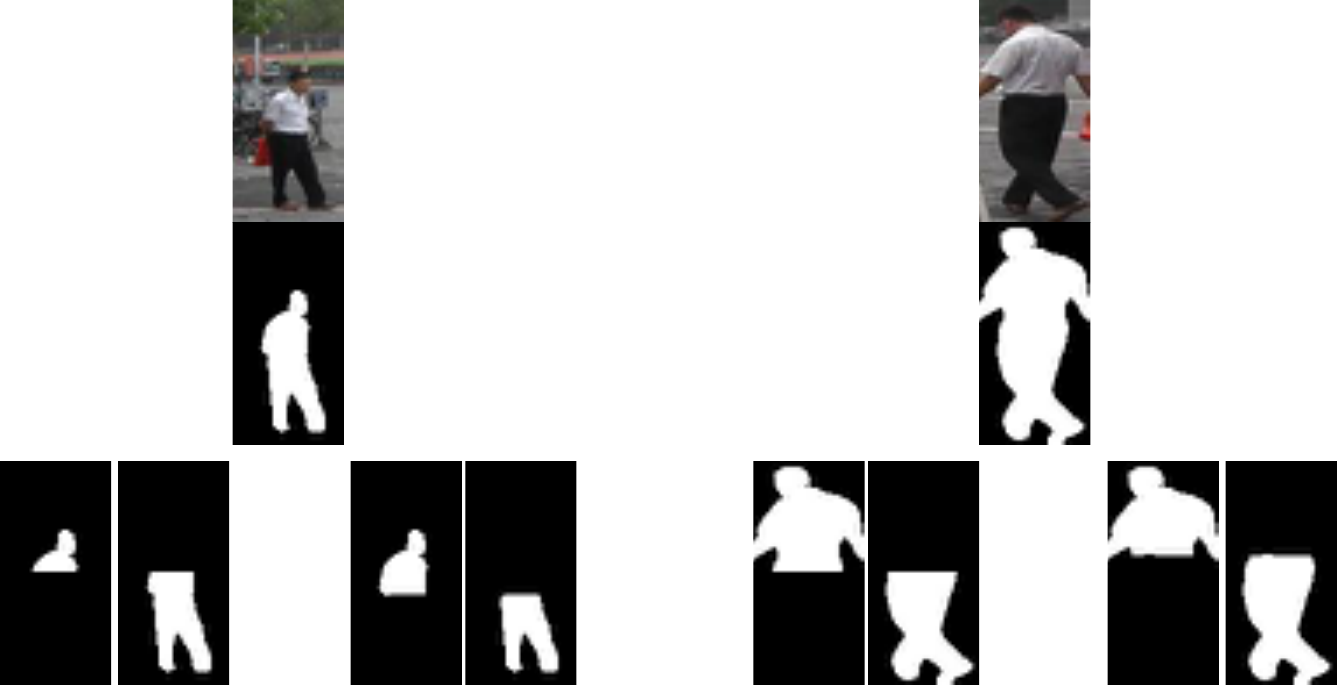}
    \caption{Split mask from the middle-line may suffer misalignment problem. The first row shows the original images, the second row shows whole-body masks. The left pair of the third row is the result of divide whole-body mask into upper-half part and bottom-half part according to the middle line. The right pair is upper-body mask and bottom-body mask.}
    \label{fig:Figure6}
\end{figure}

\begin{figure*}[ht]
    \centering
    \includegraphics[width=\textwidth]{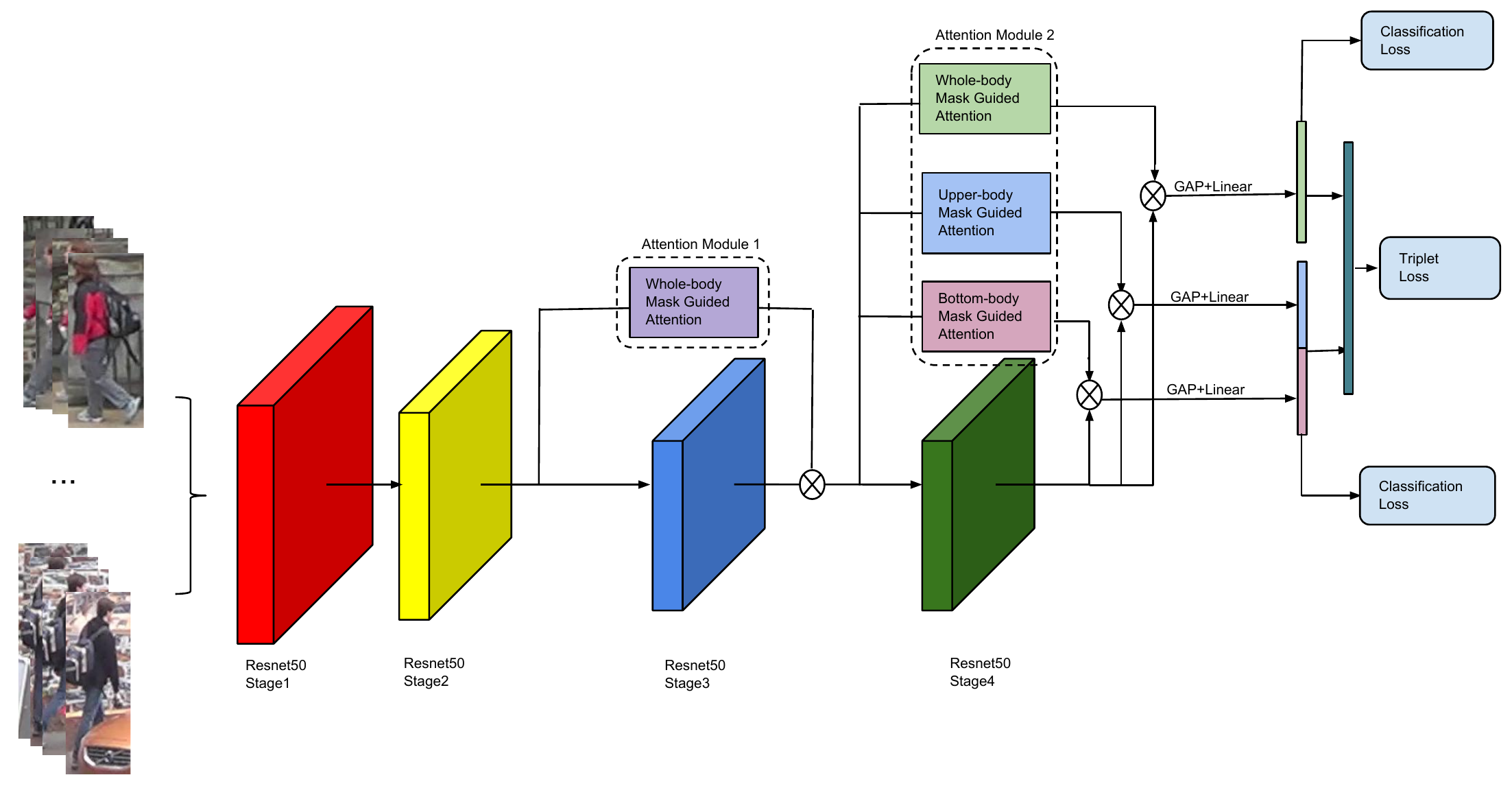}
    \caption{Structure of our proposed network. Backbone is resnet50. There are two attention modules in the network. The first attention module is guided by whole-body mask. The second attention module has three branches which are guided by whole-body mask, upper-body mask and bottom-body mask respectively.}
    \label{fig:Figure2}
\end{figure*}

\begin{figure}[ht]
    \includegraphics[width=0.5\textwidth]{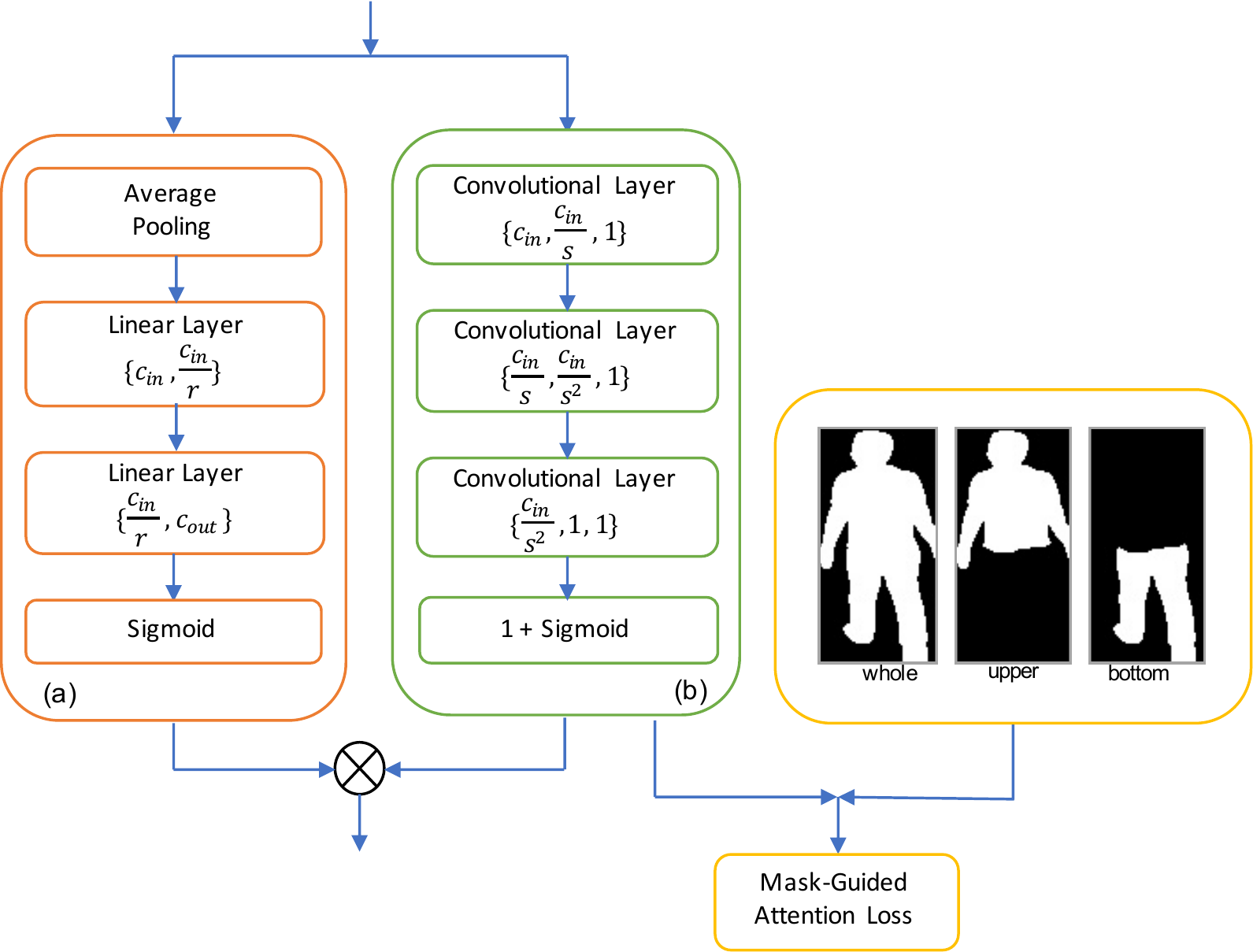}
    \caption{Structure of our mask guided attention branch. $(a)$ is the channel attention. $(b)$ is the spatial attention. Channel attention consists of one average pooling layer and two linear layers. Spatial attention is composed of three convolutional layers. Spatial attention is guided by body masks.}
    \label{fig:Figure3}
\end{figure}

In this paper, we propose a multi-scale body-part mask guided attention network. The overview of our proposed network is shown in Figure~\ref{fig:Figure2}. The detail structure of our body-part guided attention modules is shown in Figure~\ref{fig:Figure3}. Our attention modules are guided not only by whole-body masks but also by different body-part masks. In this paper, body masks are separated into upper-body masks and bottom-body masks.

There are two attention modules in our proposed network. The first attention module is guided only by whole-body masks. The second attention module has three branches. The first branch is guided by whole-body mask, the second branch is guided by upper-body mask, and the third branch is guided by bottom-body mask.

\subsection{Overall Architecture}
Variety of network structures are used in person re-identification \cite{resnet, inception, residualattention}. In our work, we employ resnet50 \cite{resnet} as backbone network with some modifications following the recent works \cite{pcb, mgn, horizontal}. The last spatial down-sampling operation is removed from resnet50. In this way, we can get a larger feature map. We also remove original global average pooling (GAP) layer and fully connected layer from resnet50. At the end of our backbone network, global average pooling layer and linear layer are employed to reduce the dimension of the output feature. The size of whole-body features is reduced to 1024, while the size of the upper-body features and the bottom-body features is reduced to 512.

\subsection{Attention Module}
Inspired by \cite{sca, hacnn}, our attention module is composed of spatial attention and channel attention. The spatial attention aims to define the importance of pixels. The channel attention is used to set weights to different channels. Body masks are used to guide the training of spatial attention as shown in Figure~\ref{fig:Figure3}. 

The input to our attention module is a 3-D tensor $\pmb{X}\in{R^{h_{in} \times w_{in} \times c_{in}}}$. The output of our attention module $\pmb{A}\in{R^{h_{out} \times w_{out} \times c_{out}}}$.

\noindent\textbf{Spatial Attention:}
Our Spatial attention consists of three convolutional layers. The input channel size of the first convolutional layer is $c_{in}$. The output channel size of it is $c_{in} / s$ where $s$ is the spatial attention reduction hyperparameter. The kernel size of the first convolutional layer is $1$. For the second convolutional layer, the input channel size is $c_{in} / s$, the output channel size is $c_{in} / s ^ 2$, and the kernel size is $1$. The input channel size of the last convolutional layer is $c_{in} / s ^ 2$, the output channel size is $1$, while the kernel size is 1. After three convolutional layers, a $Sigmoid$ function is applied to the spatial attention output. Inspired by \cite{resnet, residualattention}, we add $1.0$ to $Sigmoid$ function. The output of the spatial attention module is $\pmb{S}\in{R^{h_{out} \times w_{out} \times 1}}$. 

There is a down-sampling operation in stage3 of the backbone resnet50. An average pooling layer is applied in the first attention module between the first convolutional layer and the second convolutional layer so that the output of the first attention module has the same dimension as the output of stage3 of the backbone network. Average pooling is not applied in the second attention module.

\noindent\textbf{Channel Attention:}
Our channel attention module is composed of one average pooling layer over spatial features and two linear layers. 

We first apply an average pooling over spatial pixels in each channel as below:
\begin{equation}
    \pmb{C}_1 = 
    \frac{1}{h_{in} \times w_{in}}\sum^{h_{in}}_{i=1}\sum^{w_{in}}_{j=1}\pmb{X}_{i,j,1:c_{in}}
\end{equation}
Then two linear layers are applied to perform squeeze-expansion. The input size of the first linear layer is $c_{in}$ and the output size is $c_{in} / r$ where $r$ is the channel attention reduction hyperparameter. The input size of the second linear layer is $c_{in} / r$, while the output size of it is $c_{out}$. After these two linear layers, a $Sigmoid$ function is applied to map the channel attention to range $0$ - $1$. The channel attention is further unsqueezed to 3-dimension attention features, $\pmb{C}\in{R^{1 \times 1 \times c_{out}}}$.

\noindent\textbf{Combine Spatial-Channel Attention}:
Spatial attention and channel attention are combined by element-wise multiplication:

\begin{equation}
    \pmb{A} = \pmb{S} \times \pmb{C}
\end{equation}
where $\pmb{A}\in{R^{h_{out} \times w_{out} \times c_{out}}}$ is the output of our attention module. 

\subsection{Multi-scale Body-part Mask Guided Attention}
In this paper, we propose a multi-scale body-part mask guided attention network. Our proposed network can not only extract global informative features but also pay attention to local discriminative features. In our proposed network, whole-body attention, upper-body attention and bottom-body attention are guided by whole-body masks, upper-body masks and bottom-body masks respectively as shown in Figure~\ref{fig:Figure2}. 

Recently, a number of re-ID methods combine global features and local features to improve the accuracy of person re-identification. In order to extract local features, images or feature maps are divided into fix-height strips. But it is difficult to handle misalignment problem when inaccurate bounding-box occurs as shown in Figure~\ref{fig:Figure6}. In our proposed network, images and feature maps are not divided into fix-height strips. Instead, local attention is directly guided by local masks so that our network can accurately locate informative local part. This mechanism can significantly reduce the influence of background clutter, occlusion, inaccurate detection, etc.  Our attention module can accurately locate the informative part and thus help the network to extract whole-body features, upper-body features and bottom-body features separately.

We design two attention modules in our proposed method. The first attention module is used to filter background influence. The second attention module help the network extract whole-body features, upper-body features and bottom-body features accurately. The first attention module takes the output from stage 2 of the backbone resnet50. The output of the attention module is performed element-wise multiplication with the output features from stage 3 of the backbone resent50. The result of the element-wise multiplication is the input of the second attention module. There are 3 branches in the second attention module. The first branch is guided by whole-body masks. It acts as whole-body attention and help extract whole-body features. The second attention is upper-body attention. Its training is guided by upper-body masks to help extract upper body features. The third attention is bottom-body attention. Its training is guided by bottom-body mask to help extract bottom body features. The output of these three attention branches are performed element-wise product multiplication with the output features from stage 4 of backbone resnet50

\subsection{Loss Function}

To learn the attention map, mask guided attention loss is introduced. We also employ softmax loss for classification and hard triplet loss for metric learning. The total loss is the sum of mask guided attention loss, softmax loss and batch-hard triplet loss.

\noindent\textbf{Mask Guided Attention Loss:}
Mask is used to guide the training of spatial attention. 
To compute the attention loss, the output of spatial attention is normalized as below:
\begin{equation}
    \pmb{S}^{norm} = 
    \frac{\pmb{S} - \min(\pmb{S})}{\max(\pmb{S}) - \min(\pmb{S})}
\end{equation}
where $\pmb{S}$ represents the spatial attention and $\pmb{S}^{norm}$ is the normalized spatial attention. Then attention loss can be calculated through Root Mean Squared Error (RMSE):
\begin{equation}
    L = \sqrt{ 
    \frac{\sum^h_{i=1}\sum^w_{j=1}\lVert \pmb{M}_{i,j} - \pmb{S}^{norm}_{i,j}\rVert^2}{n_{batch}}}
\end{equation}
where $\pmb{M}$ is the resized human body mask, and $n_{batch}$ is the batch size at training stage.

The total Mask-guided attention loss is the sum of four parts.
\begin{equation}
    L_{att} = L_{att}^{1, w} + L_{att}^{2, w} + \lambda_0 L_{att}^{2, u} + \lambda_0 L_{att}^{2, b}
\end{equation}
where $L_{att}^{1, w}$ is the attention loss of the first attention module. $L_{att}^{2, w}$ is the attention loss of the first branch in the second attention module. $L_{att}^{2, u}$ is the upper-body mask guided attention loss in the second attention module. $L_{att}^{2, b}$ is the bottom-body mask guided attention loss in the second attention module. $\lambda_0$ is used to balance the whole-body mask guided attention loss and part-body mask guided attention loss.

\noindent\textbf{Softmax Loss:} At the end of our proposed network, global average pooling and linear layer are performed to reduce the dimensions of whole body features, upper-body features and bottom-body features to 1024, 512, 512 respectively. Upper-body features and bottom-body features are concatenated together to form a 1024-dimension local body-part features. Then softmax loss can be used to compute the whole body loss as well as the body-part loss  through these whole-body features and local body-part features. The loss functions are expressed as below: 

\begin{equation}
    L_{softmax}^w = -\frac{1}{n_{batch}}\sum^{n_{batch}}_{i=1}{log\frac{\exp({\pmb{W}_{y_i}^w}^{T}\pmb{f}^w_i)}{\sum^{N}_{j=1}\exp(\pmb{W}^{w}_{j}\pmb{f}^w_i)}}
\end{equation}

\begin{equation}
    L_{softmax}^l = -\frac{1}{n_{batch}}\sum^{n_{batch}}_{i=1}{log\frac{\exp({\pmb{W}_{y_i}^l}^{T}\pmb{f}^l_i)}{\sum^{N}_{j=1}\exp(\pmb{W}^{l}_{j}\pmb{f}^l_i)}}
\end{equation}

Equation (6) illustrates the whole-body softmax loss where $\pmb{f}^w$ represents 1024 dimension whole-body features and ${\pmb{W}^w_k}$ is the whole-body weight vector for identity $k$. Equation (7) illustrates the local-body softmax loss with $\pmb{f}^l$ representing the local-body features which is a concatenation of 512 dimensions of upper-body features and 512 dimensions of bottom-body features. ${\pmb{W}^l_k}$ representing the part-body weight vector for identity $k$. $N$ is the number of identities in the training dataset.  


\noindent\textbf{Triplet Loss:}
Triplet loss and its variation \cite{triplet, in_defense_triplet, Ristani_2018} is widely used in person re-identification. Batch-hard triplet loss is also applied in our proposed method to improve the accuracy.

Whole-body features, upper-body features and bottom-body features are concatenated together and normalized. Dimension of the new features is 2048. The new features $\pmb{f}^{all}$ are also used at inference stage.

The triplet loss can be formulated as below:
\begin{equation}
\begin{split}
L_{triplet} = \frac{1}{n_{batch}}\sum^{P}_{i=1}\sum^{K}_{a=1}[m &+  \max_{p=1...K}\lVert {\pmb{f}_a^{all}}^{(i)} - {\pmb{f}_p^{all}}^{(i)} \rVert_2 \\&-  \min_{\substack{n=1...K \\ j=1...P \\ j \neq i}}\lVert {\pmb{f}_a^{all}}^{(i)} - {\pmb{f}_n^{all}}^{(j)} \rVert_2]_+
\end{split}
\end{equation}
where ${\pmb{f}_a^{all}}^{(i)}$, ${\pmb{f}_p^{all}}^{(i)}$, and ${\pmb{f}_n^{all}}^{(i)}$ are the concatenated and normalized features of anchor, positive and negative samples respectively. P is the number of identities in each mini-batch and K is the number of images for each identity.

By combining all the above loss, our final loss function for the end to end multi-scale body-part mask guided attention network is as below:

\begin{equation}
    Loss = L_{softmax}^w + L_{softmax}^l + \lambda_1 L_{triplet} + \lambda_2 L_{att}
\end{equation}
where $\lambda_1$ and $\lambda_2$ are used to balance different loss. 

\noindent\textbf{}

\section{Experiments}
\begin{figure*}[ht]
    \centering
    \includegraphics[width=\textwidth]{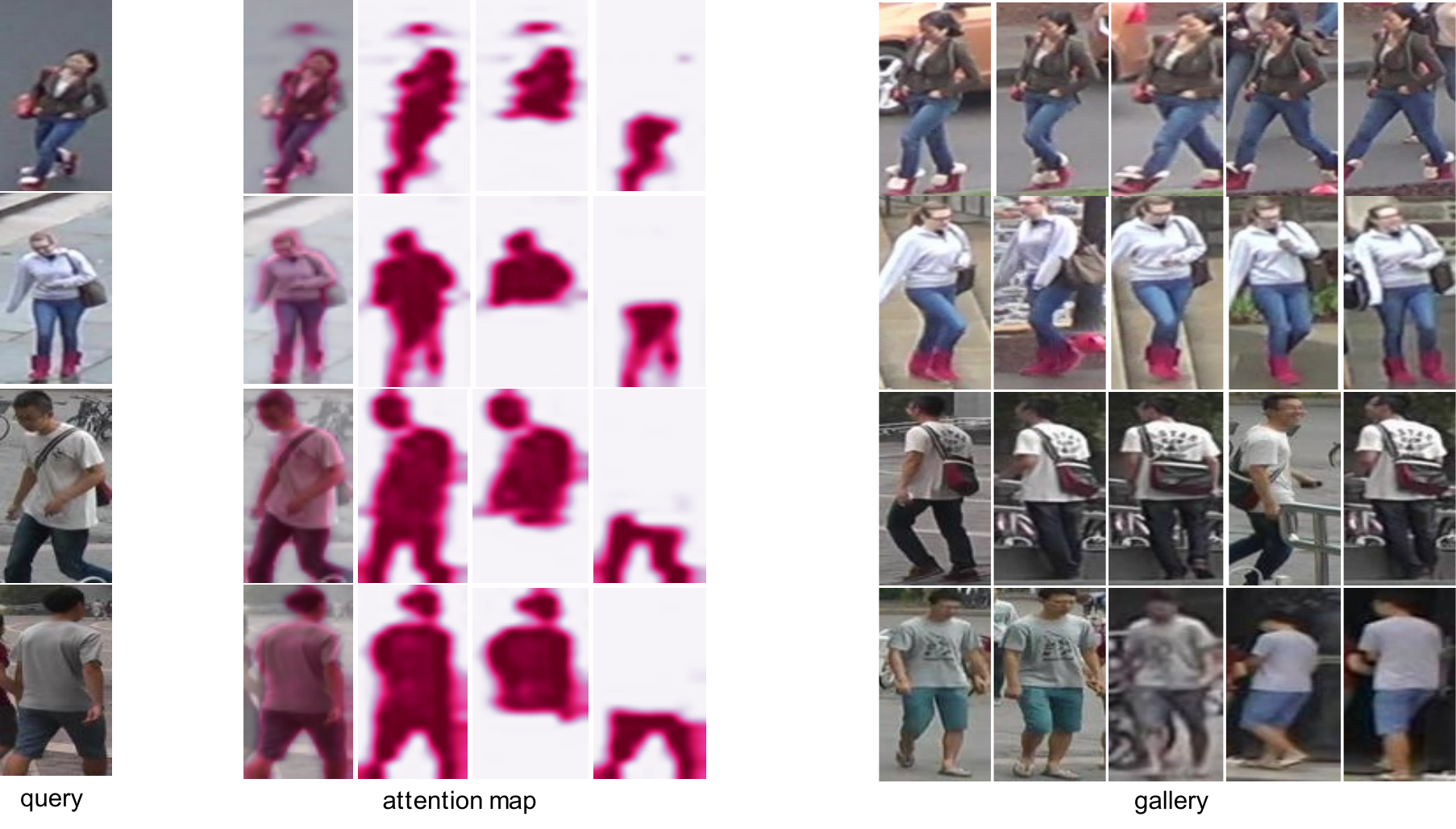}
    \caption{The top-5 ranking list for the query images on Market-1501 and DukeMTMC-reID by MMGA. The retrieved images are from the gallery set and not from the same camera as the query images. The first column is query images. Whole-body attention maps on original images, whole-body attention maps, upper-body attention maps, bottom-body attention maps from our MMGA are in the second to fifth columns. The right side shows the retrieved top-5 ranking images with our proposed MMGA method. All top-5 are correct.}
    \label{fig:Figure5}
\end{figure*}

\subsection{Dataset and Evaluation Protocol}
We choose two person re-ID benchmarks for evaluation, that is, Market-1501 \cite{market1501} and DukeMTMC-reID \cite{duke}. It is necessary to introduce these datasets and protocol we use.

 \noindent\textbf{DukeMTMC-reID:}
DukeMTMC-reID is a subset of the DukeMTMC dataset for image-based re-identification. It consists of 36411 images of 1812 identities from 8 different cameras. There are 1404 identities appearing in more than two cameras and 408 identities (distractor ID) appearing in only one camera. 16522 images of 702 persons are divided into training set. 19889 images of the remaining identities are divided into testing set, with 2228 in query set and 17661 in gallery set.

\noindent\textbf{Market-1501:}
Market-1501 contains 32668 images of 1501 identities. A total of six cameras were used, including 5 high-resolution cameras and one low-resolution camera. 12936 images of 751 identities are divided into training set. 19732 images of the remaining 750 identities are divided into testing set. 3368 images are in query set. The maximum number of query images is 6 for an identity.

\noindent\textbf{Protocols:}
We adopt Cumulative Matching Characteristic (CMC) and mean average precision (mAP) to evaluate our method. Rank-1, rank-5, rank-10 and mAP results are reported. All results reported in this paper are under single-query setting. During evaluation, following the evaluation method in MGN \cite{mgn}, we extract the features of the original images and the horizontally flipped images. The average of these features are used as the final features. Re-ranking is not used in this paper. 

\subsection{Implementation Details}
\noindent\textbf{Body Mask:}
20 different body parts (like head, hand, arm, leg, etc) can be segmented with JPPNet \cite{jppnetj, jppnetc}. According to these body part masks, whole human body is divided into upper body-part group and bottom body-part group. The reason why we only classify body parts into these two groups is that body limb masks cannot be generated so accurately with JPPNet model. Many wrong masks appear or some body-part masks are missing when upper/bottom body is further subdivided. However, if the body-part mask can be labeled by hand or generated by a more accurate model, we believe that a better performance can be achieved as attention model can focus on more local details of human body parts . 

We resize the masks to $24 \times 8$, the same size as the output of attention modules and use them to guide attention training.

\noindent\textbf{Data pre-processing:}
We follow commonly used data augmentation methods in person re-identification. All images are resized to $384 \times 128$. Horizontal random flipping and random erasing \cite{randomerase} is used in the training phase.

\noindent\textbf{Network settings:}
Resnet50 with pretrained parameters on ImageNet \cite{imagenet} is adopted as the backbone network. In our attention module, we set the spatial attention reduction hyperparameter $s$ equals 8 and the channel attention reduction hyperparameter $r$ equals 8.

\noindent\textbf{Loss:}
The margin of batch hard triplet loss is 0.3. 24 identities are sampled in each mini-batch and 4 images are sampled for each identities. The number of images in each mini-batch is 96. The loss hyperparameters are $\lambda_0 = 0.5$, $\lambda_1 = 2.0$ and $\lambda_2 = 0.1$. 

\noindent\textbf{Optimization:}
We adopt SGD optimizer with a weight decay of $5 \times 10 ^ {-4}$. The initial learning rate for the backbone network is 0.05, and the initial learning rate of other parameters is 0.1. The learning rate is decayed by a factor of 0.5 for every 90 epochs. Our model is totally trained on 900 epochs. Our model is implemented on Pytorch platform and trained with two NVIDIA 1080Ti GPUs.

\subsection{Ablation Study}
\begin{table}
  \centering
  \caption{Ablation study on Market-1501 and DukeMTMC-reID. WMGA refers to the method that attention is guided only by whole-body masks. DMGA refers to the method that divides whole-body masks from middle into upper part and bottom part and these masks are used to guide attention training. MMGA refers to our multi-scale body-part mask guided attention method.}
    \begin{tabular}{l|r|r|r|r}
    \hline
    \multicolumn{1}{c|}{ {Model}} & \multicolumn{2}{c|}{Market-1501} & \multicolumn{2}{c}{DukeMTMC-reID} \\ \cline{2-5}
          & \multicolumn{1}{c|}{Rank 1} & \multicolumn{1}{c|}{mAP} & \multicolumn{1}{c|}{Rank 1} & \multicolumn{1}{c}{mAP} \\
    \hline
    Baseline(\%) & 92.7 & 83.5  & 86.9  & 73.9 \\
    Baseline+Att(\%) & 93.4 & 85.0 & 87.5 & 75.9 \\
    WMGA(\%)  & 94.3 & 86.4 & 88.3 & 77.3 \\
    DMGA(\%) & 94.4 & 86.9  & 88.6 & 77.8 \\
    MMGA(\%) & 95.0 & 87.2 & 89.5 & 78.1 \\
    \hline
    \end{tabular}%
  \label{tab:abalation}%
\end{table}%

\begin{table*}[htbp]
  \centering
  \caption{Comparison of results on Market-1501 and DukeMTMC-reID.}
    \begin{tabular}{l|r|r|r|r|r|r}
    \hline
    \multicolumn{1}{c|}{{Model}} & \multicolumn{4}{c|}{Market-1501} & \multicolumn{2}{c}{DukeMTMC-reID} \\
\cline{2-7}          & \multicolumn{1}{l|}{Rank 1} & \multicolumn{1}{l|}{Rank 5} & \multicolumn{1}{l|}{Rank 10} & \multicolumn{1}{l|}{mAP} & \multicolumn{1}{l|}{Rank 1} & \multicolumn{1}{l}{mAP} \\
    \hline
    BoW+kissme \cite{market1501} &    44.4   &     63.9  &72.2       &     20.8  &    25.1   & 12.2 \\
    WARCA \cite{warca} &   45.2    & 68.1      & 76.0      & --      &    --   & -- \\
    SVDNet \cite{svdnet} &    82.3   &    92.3   &    95.2   &  62.1     &    76.7   & 56.8 \\
    PAN \cite{pan}   &    82.8   &  --     &   --    &    63.4   &   71.6    & 51.5 \\
    PAR \cite{par}   &   81.0    &   92.0    &    94.7   &   63.4    &   --    & -- \\
    MultiLoss \cite{multiloss} &   83.9    &    --   &   --    &    64.4   &   --    & -- \\
    TripletLoss \cite{in_defense_triplet} &   84.9    &   94.2    &    --   &    69.1   &   --    &  --\\
    MultiScale \cite{multiscale} &   88.9    &   --    &     --  &   73.1    &   79.2    & 60.6 \\
    MLFN \cite{mlfn}  &    90.0   &   --    &    --   &   74.3    &    81.0   & 62.8 \\
    HA-CNN \cite{hacnn} &   91.2    &    --   &   --    &  75.7     &    80.5   &  63.8\\
    AACN \cite{aacn} &    85.9   &    --   &   --    &   66.9    &    76.8   &  59.3\\
    MSCAN \cite{maskguided} &    83.8   &    --   &   --    &   74.3    &    --   &  --\\
    AlignedReID \cite{alignedreid} &   91.0    &    96.3   &  --     &   79.4    &    --   & -- \\
    Deep-Person \cite{deepperson} &    92.3   &    --   &   --    &   79.5    &    80.9   &  64.8\\
    PCB+RPP \cite{pcb} &    93.8   &   97.5    &   98.5    &  81.6     &  83.3     & 69.2 \\
    HPM \cite{horizontal}   &    94.2   &   97.5    &   98.5    &   82.7    &   86.6    & 74.3 \\
    Attention-Driven \cite{attentiondriven} &   95.0    &    \pmb{98.3}   &  \pmb{99.1}     &   86.5    &    86.4   & 74.6 \\
    MGN \cite{mgn}   &    \pmb{95.7}   &    --   &    --   &    86.9   &  88.7     & \pmb{78.4} \\
    \hline
    MMGA (Ours) &   95.0    & 98.2   &   98.9    &   \pmb{87.2}    &   \pmb{89.5}    & 78.1 \\
    \hline
    \end{tabular}%
  \label{tab:comparison}%
\end{table*}%

We perform comprehensive ablation studies on Market-1501 and DukeMTMC-reID. 

\noindent\textbf{Baseline:}
In Table~\ref{tab:abalation}, Baseline is trained only on backbone resnet50 without attention module. After the backbone resnet50, a global average pooling layer and a linear layer is applied. A 1024-dimension global features are extracted. Classification loss and triplet loss are combined to train the model. The hyperparameter for balancing the weight of triplet loss and classification loss is the same as mentioned in implementation details section. Rank-1/mAP of baseline is 92.7\%/83.5\% on Market-1501 dataset and 86.9\%/73.9\% on DukeMTMC-reID dataset.

\noindent\textbf{Baseline+Att:}
In Table~\ref{tab:abalation}, Baseline+Att denotes the model that two attention modules are applied to the baseline. The attention modules are not guided by body masks in training stage and there is only whole-body attention branch in the second attention module. Results are listed in Table~\ref{tab:abalation}. On Market-1501 dataset, rank-1/mAP is 93.4\%/85.0\% and on DukeMTMC-reID dataset, rank-1/mAP is 87.5\%/75.9\%. Attention mechanism can improve the accuracy by +0.7\%/+1.5\% for rank1/mAP on Market-1501 dataset and +0.6\%/+2.0\% on DukeMTMC-reID dataset. 

\noindent\textbf{Whole-body Mask-guided attention (WMGA):}
We evaluate the effect of mask-guided attention mechanism. WMGA refers to the method that attention modules are guided by whole-body masks. Part-body masks are not used in WMGA. WMGA achieves rank-1/mAP=94.3\%/86.4\% on Market-1501 and rank-1/mAP=88.3\%/77.3\% on DukeMTMC-reID. WMGA improves the accuracy of Baseline+Att by a large margin. The experiments illustrate that using masks to guide the training of attention can significantly improve the accuracy of person re-identification.

\noindent\textbf{Multi-scale Body-part Mask guided Attention (MMGA) versus WMGA:}
 In our proposed method (MMGA), not only whole-body masks are used but also part-body masks are used to guide the training of attention modules. Thanks to MMGA method, our network can predict in detail where the person is and locate upper-body and bottom-body accurately. This mechanism can significantly reduce the impact of background clutter, varient of pose and misalignment. The results of MMGA are listed in Table~\ref{tab:abalation}. MMGA achieves 95.0\%/87.2\% in rank-1/mAP on Market-1501 outperforming WMGA by +0.7\%/+0.8\%. On DukeMTMC-reID, MMGA achieves 89.5\%/78.1\% in rank-1/mAP improving the accuracy of WMGA by +1.2\%/+0.8\%.

\noindent\textbf{Ours (MMGA) versus Divided-Mask guided attention (DMGA):}
We also divided whole-body masks into upper-half parts and bottom-half parts according to the middle line and use this kind of masks to guide the training of our attention module (DMGA). The upper-half masks and bottom-half masks can somewhat act as upper-body masks and bottom-body masks. However, dividing masks from the middle can not locate the upper-body part and bottom-body part accurately, especially for misalignment images. DMGA achieves 94.4\%/86.9\% in rank-1/mAP on Market-1501 and 88.6\%/77.8\% on DukeMTMC-reID. Experiments show that the results of DMGA are less than MMGA by -0.6\%/-0.3\% on Market-1501 and -0.9\%/-0.3\% on DukeMTMC-reID. 

The above ablation study illustrates that attention mechanism can help feature extraction of person re-identification and thus using multi-scale body-part mask to guide attention training can significantly improve the accuracy of person re-identification.

In Figure~\ref{fig:Figure5}, we list four inaccurate person detection images from query set on Market-1501 and DukeMTMC-reID. Our proposed method (MMGA) can locate where the informative regions are and learn upper-body attention as well as bottom-body attention correctly. Attention maps of MMGA network are shown in Figure~\ref{fig:Figure5}. The top-5 retrieve results for these four inaccurate detection images are all correct.

\subsection{Comparison with State-of-the-Art}

We compare our proposed multi-scale body-part mask guided attention method with current state-of-the-art methods on Market-1501 and DukeMTMC-reID. 

Strip-based methods \cite{mgn, pcb, horizontal} divide features into several horizontal stripes to extract local features but it is difficult to handle misalignment problem. \cite{attentiondriven} uses key-points to divide original images into several parts so that they can exploit global features and local features. Attention-based methods \cite{hacnn, attentiondriven} use attention mechanism to handle background clutter and extract important features. \cite{maskguided} add masks into input to form four-channeled inputs which is incompatible with the pre-trained resnet models. Our proposed multi-scale part-body mask guided attention method uses multi-scale masks to guide the training of different attention modules. Our mothod can locate the important location and extract global features and local features accurately. Our methods achieves the state-of-art performance. Results are shown in Table~\ref{tab:comparison}.

\noindent\textbf{Market-1501:} 
On Market-1501 dataset, our proposed MMGA method outperforms all the state-of-the-art algorithms and exceeds the current best model MGN by +0.3\% in mAP. Our method achieves the second best result in rank-1. 

\noindent\textbf{DukeMTMC-reID:} 
Our proposed MMGA achieves 89.5\%/78.1\% in rank-1/mAP on DukeMTMC-reID. Though we achieve the second best result in mAP, 0.3\% lower than MGN, in rank-1, our method outperforms all other models and exceeds the state-of-the-art method MGN by +0.8\%. We should pay attention that our model is smaller than MGN. In MGN, backbone resnet50 is divided into three branches at stage 3 and these three branches do not share weights with each other. 

\section{Conclusion and Discussion}
In this paper, we propose a multi-scale body-part mask guided attention network. The training of attention modules is guided by whole-body masks, upper-body masks and bottom-body masks. Our method can accurately locate the important parts of human body. Experiments show that our proposed method can significantly improve the accuracy of person re-identification and our method achieves the state-of-the-art result. What's more, in this paper, we only divided body masks into upper-body masks and bottom-body masks. We believe if we can divide masks into finer details, we can further improve the accuracy of person re-identification. We leave this experiment as future work.

{\small 
\bibliographystyle{ieee_fullname}
\bibliography{egbib}

\begin{thebibliography}{10}\itemsep=-1pt

\bibitem{deepperson}
Xiang Bai, Mingkun Yang, Tengteng Huang, Zhiyong Dou, Rui Yu, and Yongchao Xu.
\newblock Deep-person: Learning discriminative deep features for person
  re-identification, 2017.

\bibitem{mlfn}
Xiaobin Chang, Timothy~M. Hospedales, and Tao Xiang.
\newblock Multi-level factorisation net for person re-identification.
\newblock {\em 2018 IEEE/CVF Conference on Computer Vision and Pattern
  Recognition}, Jun 2018.

\bibitem{Chen2018PersonSV}
Di Chen, Shanshan Zhang, Wanli Ouyang, Jian~Xi Yang, and Ying Tai.
\newblock Person search via a mask-guided two-stream cnn model.
\newblock In {\em ECCV}, 2018.

\bibitem{sca}
Long Chen, Hanwang Zhang, Jun Xiao, Liqiang Nie, Jian Shao, Wei Liu, and
  Tat-Seng Chua.
\newblock Sca-cnn: Spatial and channel-wise attention in convolutional networks
  for image captioning.
\newblock In {\em Proceedings of the IEEE Conference on Computer Vision and
  Pattern Recognition}, pages 5659--5667, 2017.

\bibitem{multiscale}
Y. {Chen}, X. {Zhu}, and S. {Gong}.
\newblock Person re-identification by deep learning multi-scale
  representations.
\newblock In {\em 2017 IEEE International Conference on Computer Vision
  Workshops (ICCVW)}, pages 2590--2600, Oct 2017.

\bibitem{cheng2016person}
De Cheng, Yihong Gong, Sanping Zhou, Jinjun Wang, and Nanning Zheng.
\newblock Person re-identification by multi-channel parts-based cnn with
  improved triplet loss function.
\newblock In {\em Proceedings of the IEEE Conference on Computer Vision and
  Pattern Recognition}, pages 1335--1344, 2016.

\bibitem{imagenet}
J. Deng, W. Dong, R. Socher, L.-J. Li, K. Li, and L. Fei-Fei.
\newblock {ImageNet: A Large-Scale Hierarchical Image Database}.
\newblock In {\em CVPR09}, 2009.

\bibitem{farenzena2010person}
Michela Farenzena, Loris Bazzani, Alessandro Perina, Vittorio Murino, and Marco
  Cristani.
\newblock Person re-identification by symmetry-driven accumulation of local
  features.
\newblock In {\em 2010 IEEE Computer Society Conference on Computer Vision and
  Pattern Recognition}, pages 2360--2367. IEEE, 2010.

\bibitem{horizontal}
Yang Fu, Yunchao Wei, Yuqian Zhou, Honghui Shi, Gao Huang, Xinchao Wang,
  Zhiqiang Yao, and Thomas Huang.
\newblock Horizontal pyramid matching for person re-identification.
\newblock {\em arXiv preprint arXiv:1804.05275}, 2018.

\bibitem{jppnetc}
Ke Gong, Xiaodan Liang, Dongyu Zhang, Xiaohui Shen, and Liang Lin.
\newblock Look into person: Self-supervised structure-sensitive learning and a
  new benchmark for human parsing.
\newblock In {\em The IEEE Conference on Computer Vision and Pattern
  Recognition (CVPR)}, July 2017.

\bibitem{densepose}
Riza~Alp Guler, George Trigeorgis, Epameinondas Antonakos, Patrick Snape,
  Stefanos Zafeiriou, and Iasonas Kokkinos.
\newblock Densereg: Fully convolutional dense shape regression in-the-wild.
\newblock {\em 2017 IEEE Conference on Computer Vision and Pattern Recognition
  (CVPR)}, Jul 2017.

\bibitem{hamdoun2008person}
Omar Hamdoun, Fabien Moutarde, Bogdan Stanciulescu, and Bruno Steux.
\newblock Person re-identification in multi-camera system by signature based on
  interest point descriptors collected on short video sequences.
\newblock In {\em 2008 Second ACM/IEEE International Conference on Distributed
  Smart Cameras}, pages 1--6. IEEE, 2008.

\bibitem{He2017MaskR}
Kaiming He, Georgia Gkioxari, Piotr Doll{\'a}r, and Ross~B. Girshick.
\newblock Mask r-cnn.
\newblock pages 2980--2988, 2017.

\bibitem{resnet}
Kaiming He, Xiangyu Zhang, Shaoqing Ren, and Jian Sun.
\newblock Deep residual learning for image recognition.
\newblock {\em 2016 IEEE Conference on Computer Vision and Pattern Recognition
  (CVPR)}, Jun 2016.

\bibitem{in_defense_triplet}
Alexander Hermans, Lucas Beyer, and Bastian Leibe.
\newblock In defense of the triplet loss for person re-identification.
\newblock {\em arXiv preprint arXiv:1703.07737}, 2017.

\bibitem{warca}
Cijo Jose and François Fleuret.
\newblock Scalable metric learning via weighted approximate rank component
  analysis.
\newblock {\em Lecture Notes in Computer Science}, page 875–890, 2016.

\bibitem{spreid}
Mahdi~M. Kalayeh, Emrah Basaran, Muhittin Gokmen, Mustafa~E. Kamasak, and
  Mubarak Shah.
\newblock Human semantic parsing for person re-identification.
\newblock {\em 2018 IEEE/CVF Conference on Computer Vision and Pattern
  Recognition}, Jun 2018.

\bibitem{contextaware}
Dangwei Li, Xiaotang Chen, Zhang Zhang, and Kaiqi Huang.
\newblock Learning deep context-aware features over body and latent parts for
  person re-identification.
\newblock {\em 2017 IEEE Conference on Computer Vision and Pattern Recognition
  (CVPR)}, Jul 2017.

\bibitem{diversity}
Shuang Li, Slawomir Bak, Peter Carr, and Xiaogang Wang.
\newblock Diversity regularized spatiotemporal attention for video-based person
  re-identification.
\newblock {\em 2018 IEEE/CVF Conference on Computer Vision and Pattern
  Recognition}, Jun 2018.

\bibitem{deepreid}
W. {Li}, R. {Zhao}, T. {Xiao}, and X. {Wang}.
\newblock Deepreid: Deep filter pairing neural network for person
  re-identification.
\newblock In {\em 2014 IEEE Conference on Computer Vision and Pattern
  Recognition}, pages 152--159, June 2014.

\bibitem{multiloss}
Wei Li, Xiatian Zhu, and Shaogang Gong.
\newblock Person re-identification by deep joint learning of multi-loss
  classification.
\newblock {\em Proceedings of the Twenty-Sixth International Joint Conference
  on Artificial Intelligence}, Aug 2017.

\bibitem{hacnn}
Wei Li, Xiatian Zhu, and Shaogang Gong.
\newblock Harmonious attention network for person re-identification.
\newblock In {\em Proceedings of the IEEE Conference on Computer Vision and
  Pattern Recognition}, pages 2285--2294, 2018.

\bibitem{jppnetj}
Xiaodan Liang, Ke Gong, Xiaohui Shen, and Liang Lin.
\newblock Look into person: Joint body parsing \& pose estimation network and a
  new benchmark.
\newblock {\em IEEE Transactions on Pattern Analysis and Machine Intelligence},
  2018.

\bibitem{lin2017improving}
Yutian Lin, Liang Zheng, Zhedong Zheng, Yu Wu, and Yi Yang.
\newblock Improving person re-identification by attribute and identity
  learning.
\newblock {\em arXiv preprint arXiv:1703.07220}, 2017.

\bibitem{Liu2017HydraPlusNetAD}
Xihui Liu, Haiyu Zhao, Maoqing Tian, Lu Sheng, Jing Shao, Shuai Yi, Junjie Yan,
  and Xiaogang Wang.
\newblock Hydraplus-net: Attentive deep features for pedestrian analysis.
\newblock {\em 2017 IEEE International Conference on Computer Vision (ICCV)},
  pages 350--359, 2017.

\bibitem{matsukawa2016person}
Tetsu Matsukawa and Einoshin Suzuki.
\newblock Person re-identification using cnn features learned from combination
  of attributes.
\newblock In {\em 2016 23rd International Conference on Pattern Recognition
  (ICPR)}, pages 2428--2433. IEEE, 2016.

\bibitem{Ristani_2018}
Ergys Ristani and Carlo Tomasi.
\newblock Features for multi-target multi-camera tracking and
  re-identification.
\newblock {\em 2018 IEEE/CVF Conference on Computer Vision and Pattern
  Recognition}, Jun 2018.

\bibitem{triplet}
Florian Schroff, Dmitry Kalenichenko, and James Philbin.
\newblock Facenet: A unified embedding for face recognition and clustering.
\newblock {\em 2015 IEEE Conference on Computer Vision and Pattern Recognition
  (CVPR)}, Jun 2015.

\bibitem{maskguided}
Chunfeng Song, Yan Huang, Wanli Ouyang, and Liang Wang.
\newblock Mask-guided contrastive attention model for person re-identification.
\newblock In {\em The IEEE Conference on Computer Vision and Pattern
  Recognition (CVPR)}, June 2018.

\bibitem{svdnet}
Yifan Sun, Liang Zheng, Weijian Deng, and Shengjin Wang.
\newblock Svdnet for pedestrian retrieval.
\newblock {\em 2017 IEEE International Conference on Computer Vision (ICCV)},
  Oct 2017.

\bibitem{pcb}
Yifan Sun, Liang Zheng, Yi Yang, Qi Tian, and Shengjin Wang.
\newblock Beyond part models: Person retrieval with refined part pooling (and a
  strong convolutional baseline).
\newblock In {\em Proceedings of the European Conference on Computer Vision
  (ECCV)}, pages 480--496, 2018.

\bibitem{inception}
Christian Szegedy, Wei Liu, Yangqing Jia, Pierre Sermanet, Scott Reed, Dragomir
  Anguelov, Dumitru Erhan, Vincent Vanhoucke, and Andrew Rabinovich.
\newblock Going deeper with convolutions.
\newblock {\em 2015 IEEE Conference on Computer Vision and Pattern Recognition
  (CVPR)}, Jun 2015.

\bibitem{varior2016gated}
Rahul~Rama Varior, Mrinal Haloi, and Gang Wang.
\newblock Gated siamese convolutional neural network architecture for human
  re-identification.
\newblock In {\em European Conference on Computer Vision}, pages 791--808.
  Springer, 2016.

\bibitem{varior2016siamese}
Rahul~Rama Varior, Bing Shuai, Jiwen Lu, Dong Xu, and Gang Wang.
\newblock A siamese long short-term memory architecture for human
  re-identification.
\newblock In {\em European Conference on Computer Vision}, pages 135--153.
  Springer, 2016.

\bibitem{mancs}
Cheng Wang, Qian Zhang, Chang Huang, Wenyu Liu, and Xinggang Wang.
\newblock Mancs: A multi-task attentional network with curriculum sampling for
  person re-identification.
\newblock In {\em ECCV}, 2018.

\bibitem{residualattention}
Fei Wang, Mengqing Jiang, Chen Qian, Shuo Yang, Cheng Li, Honggang Zhang,
  Xiaogang Wang, and Xiaoou Tang.
\newblock Residual attention network for image classification.
\newblock {\em 2017 IEEE Conference on Computer Vision and Pattern Recognition
  (CVPR)}, Jul 2017.

\bibitem{wang2016joint}
Faqiang Wang, Wangmeng Zuo, Liang Lin, David Zhang, and Lei Zhang.
\newblock Joint learning of single-image and cross-image representations for
  person re-identification.
\newblock In {\em Proceedings of the IEEE Conference on Computer Vision and
  Pattern Recognition}, pages 1288--1296, 2016.

\bibitem{mgn}
Guanshuo Wang, Yufeng Yuan, Xiong Chen, Jiwei Li, and Xi Zhou.
\newblock Learning discriminative features with multiple granularities for
  person re-identification.
\newblock In {\em Proceedings of the 26th ACM International Conference on
  Multimedia}, MM '18, pages 274--282, New York, NY, USA, 2018. ACM.

\bibitem{nonlocal}
Xiaolong Wang, Ross Girshick, Abhinav Gupta, and Kaiming He.
\newblock Non-local neural networks.
\newblock {\em 2018 IEEE/CVF Conference on Computer Vision and Pattern
  Recognition}, Jun 2018.

\bibitem{xiao2016learning}
Tong Xiao, Hongsheng Li, Wanli Ouyang, and Xiaogang Wang.
\newblock Learning deep feature representations with domain guided dropout for
  person re-identification.
\newblock In {\em Proceedings of the IEEE conference on computer vision and
  pattern recognition}, pages 1249--1258, 2016.

\bibitem{aacn}
Jing Xu, Rui Zhao, Feng Zhu, Huaming Wang, and Wanli Ouyang.
\newblock Attention-aware compositional network for person re-identification.
\newblock {\em 2018 IEEE/CVF Conference on Computer Vision and Pattern
  Recognition}, Jun 2018.

\bibitem{attentiondriven}
Fan Yang, Ke Yan, Shijian Lu, Huizhu Jia, Xiaodong Xie, and Wen Gao.
\newblock Attention driven person re-identification.
\newblock {\em Pattern Recognition}, 86:143 -- 155, 2019.

\bibitem{Yao2019DeepRL}
Hantao Yao, Shiliang Zhang, Yongdong Zhang, Jintao Li, and Qi Tian.
\newblock Deep representation learning with part loss for person
  re-identification.
\newblock {\em IEEE transactions on image processing : a publication of the
  IEEE Signal Processing Society}, 2019.

\bibitem{selfattention}
Han Zhang, Ian~J. Goodfellow, Dimitris~N. Metaxas, and Augustus Odena.
\newblock Self-attention generative adversarial networks.
\newblock {\em arXiv:1805.08318}, 2018.

\bibitem{alignedreid}
Xuan Zhang, Hao Luo, Xing Fan, Weilai Xiang, Yixiao Sun, Qiqi Xiao, Wei Jiang,
  Chi Zhang, and Jian Sun.
\newblock Alignedreid: Surpassing human-level performance in person
  re-identification.
\newblock {\em arXiv preprint arXiv:1711.08184}, 2017.

\bibitem{par}
Liming Zhao, Xi Li, Yueting Zhuang, and Jingdong Wang.
\newblock Deeply-learned part-aligned representations for person
  re-identification.
\newblock {\em 2017 IEEE International Conference on Computer Vision (ICCV)},
  Oct 2017.

\bibitem{market1501}
L. {Zheng}, L. {Shen}, L. {Tian}, S. {Wang}, J. {Wang}, and Q. {Tian}.
\newblock Scalable person re-identification: A benchmark.
\newblock In {\em 2015 IEEE International Conference on Computer Vision
  (ICCV)}, pages 1116--1124, Dec 2015.

\bibitem{duke}
Zhedong Zheng, Liang Zheng, and Yi Yang.
\newblock Unlabeled samples generated by gan improve the person
  re-identification baseline in vitro.
\newblock In {\em Proceedings of the IEEE International Conference on Computer
  Vision}, 2017.

\bibitem{pan}
Zhedong Zheng, Liang Zheng, and Yi Yang.
\newblock Pedestrian alignment network for large-scale person
  re-identification.
\newblock {\em IEEE Transactions on Circuits and Systems for Video Technology},
  page 1–1, 2018.

\bibitem{randomerase}
Zhun Zhong, Liang Zheng, Guoliang Kang, Shaozi Li, and Yi Yang.
\newblock Random erasing data augmentation, 2017.

\end{thebibliography}
}

\end{document}